\documentclass[runningheads]{llncs}
\usepackage[T1]{fontenc}
\usepackage{graphicx,verbatim}
\usepackage{amsfonts,amssymb}
\usepackage{amsmath,multirow,makecell}

\begin{document}
\newcommand{\methodfullnameA}{Temporal Asymmetric Feature Propagation Network}
\newcommand{\methodnamea}{TAFPNet}

\newcommand{\methodfullnameB}{Temporal Query Propagator}
\newcommand{\methodnameb}{TQP}

\newcommand{\methodfullnameC}{Aggregated Asymmetric Feature Pyramid}
\newcommand{\methodnamec}{AAFP}
\title{Temporal Propagation of Asymmetric Feature Pyramid for Surgical Scene Segmentation}

\author{Cheng Yuan$^{1}$, Yutong Ban$^{1}$}  
\authorrunning{Anonymized Author et al.}
\institute{$^{1}$UM-SJTU Joint Institute, Shanghai Jiao Tong University, Shanghai, China \\
    \email{yban@sjtu.edu.cn}}

\maketitle              
\begin{abstract}
Surgical scene segmentation is crucial for robot-assisted laparoscopic surgery understanding. Current approaches face two challenges: (i) static image limitations including ambiguous local feature similarities and fine-grained structural details, and (ii) dynamic video complexities arising from rapid instrument motion and persistent visual occlusions. While existing methods mainly focus on spatial feature extraction, they fundamentally overlook temporal dependencies in surgical video streams. To address this, we present temporal asymmetric feature propagation network, a bidirectional attention architecture enabling cross-frame feature propagation. The proposed method contains a temporal query propagator that integrates multi-directional consistency constraints to enhance frame-specific feature representation, and an aggregated asymmetric feature pyramid module that preserves discriminative features for anatomical structures and surgical instruments. Our framework uniquely enables both temporal guidance and contextual reasoning for surgical scene understanding. Comprehensive evaluations on two public benchmarks show the proposed method outperforms the current SOTA methods by a large margin, with +16.4\% mIoU on EndoVis2018 and +3.3\% mAP on Endoscapes2023. The code will be publicly available after paper acceptance.



\end{abstract}
\section{Introduction}

\begin{figure}
\includegraphics[width=\textwidth]{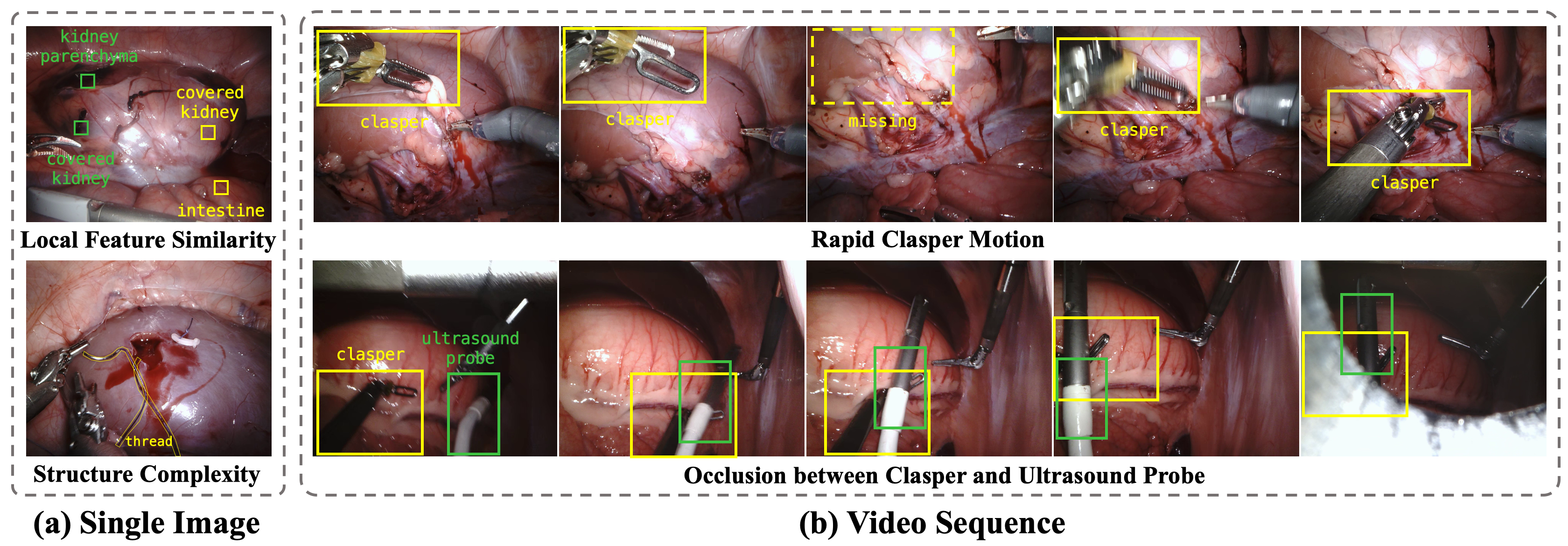}
\caption{Challenges in surgical scene segmentation: (a) local feature similarity and fine-grained structure complexity in a single image; (b) rapid object motion and inevitable interaction occlusion in video sequences.} \label{fig1}
\end{figure}

The advancement of robot-assisted minimally invasive laparoscopic surgery has heightened the importance of precise surgical scene segmentation, which serves as a fundamental prerequisite for subsequent understanding tasks including pose estimation~\cite{ref1}, triplet recognition~\cite{ref2}, and safety assessment~\cite{ref3}. In addition, separately for augmented reality and automatic annotation, referring to the scene segmentation mask can reduce render errors and labor consumption. However, achieving accurate scene segmentation is challenging, due to the local feature similarity among various anatomies and the fine-grained structure complexity of deformable instruments, as shown in Fig.~\ref{fig1}(a). Moreover, the instance recognition performance further degrades because of blur from rapid instrument manipulation and inevitable interaction occlusion, as shown in Fig.~\ref{fig1}(b).

To overcome these challenges, existing methods have evolved along two main methodological paradigms: (1) The first category employs specialized convolution networks that maximize spatial information modeling through hierarchical feature extraction. Yang et al.~\cite{ref4} implemented atrous spatial pyramid to capture multiscale context representations. Ni et al.~\cite{ref5} proposed a space-squeeze reasoning network incorporating low-rank bilinear feature fusion for different surgical region segmentation. Recent innovation by Liu et al.~\cite{ref6} introduced a large kernel attention strategy to long-strip targets, establishing superior results in instrument-tissue segmentation. (2) The second category capitalizes the transformer architecture, adapting its attention mechanism to get global information perception in segmentation. Groundbreaking work by Carion et al.~\cite{ref7} established an end-to-end query-based detection framework through bipartite matching loss. Pioneering segmentation methods like Mask2Former~\cite{ref9} and MaskDINO~\cite{ref10} demonstrated the effectiveness and versatility of query-based dense predictions. These methods achieved great performance in the natural image field, while they generally disable when meeting specific challenges of surgical videos. 

Surgical videos contain more plentiful information compared to a single static image, especially temporal clues. Existing methods proved the effectiveness of temporal consistency in several surgical vision tasks, such as phase recognition~\cite{ref12,ref13} and instrument tracking~\cite{ref14,ref15}. However, only a limited number of temporal networks on pixel-level dense prediction appeared recently. Jin et al.~\cite{ref8} extended the transformer architecture through intra- and inter-video relation modeling to capture global dependencies. It focused on tackling the class imbalance problem and produced relatively coarse segmentation. Thus, how to construct temporal propagation and incorporate it into feature enhancement is essential to achieve superior segmentation performance.

We present Temporal Asymmetric Feature Propagation Network (TAFPNet) for surgical video segmentation, addressing spatial ambiguity between anatomy and instruments. It incorporates two perception enhancement modules into a bidirectional attention network architecture. Temporal coherence is maintained via dynamic query propagation, forming occlusion-resistant feature tubes. This integration of structural asymmetry and temporal continuity enables accurate segmentation of both anatomical tissue and dynamic instruments in challenging laparoscopic scenarios. Our main contributions are summarized as follows:

\begin{enumerate}
    \item We propose TAFPNet, a novel surgical scene segmentation framework that combines temporal dynamics with structural asymmetry.
    \item The designed temporal query propagator integrated temporal dependency into the transformer encoding. Meanwhile, the aggregated asymmetric feature pyramid integrates the disentangled instrument-tissue features along the time dimension.
    \item The proposed TAFPNet achieves the best overall performance by +16.4\% mIoU on EndoVis2018~\cite{ref16} and +3.3\% mAP on Endoscapes2023~\cite{ref17}. It outperforms in granular structure segmentation with 14.8\% and 4.5\% of improvement on hepatocystic triangle and cystic artery, respectively.
\end{enumerate}

\section{Method}
\begin{figure} [t]
\includegraphics[width=\textwidth]{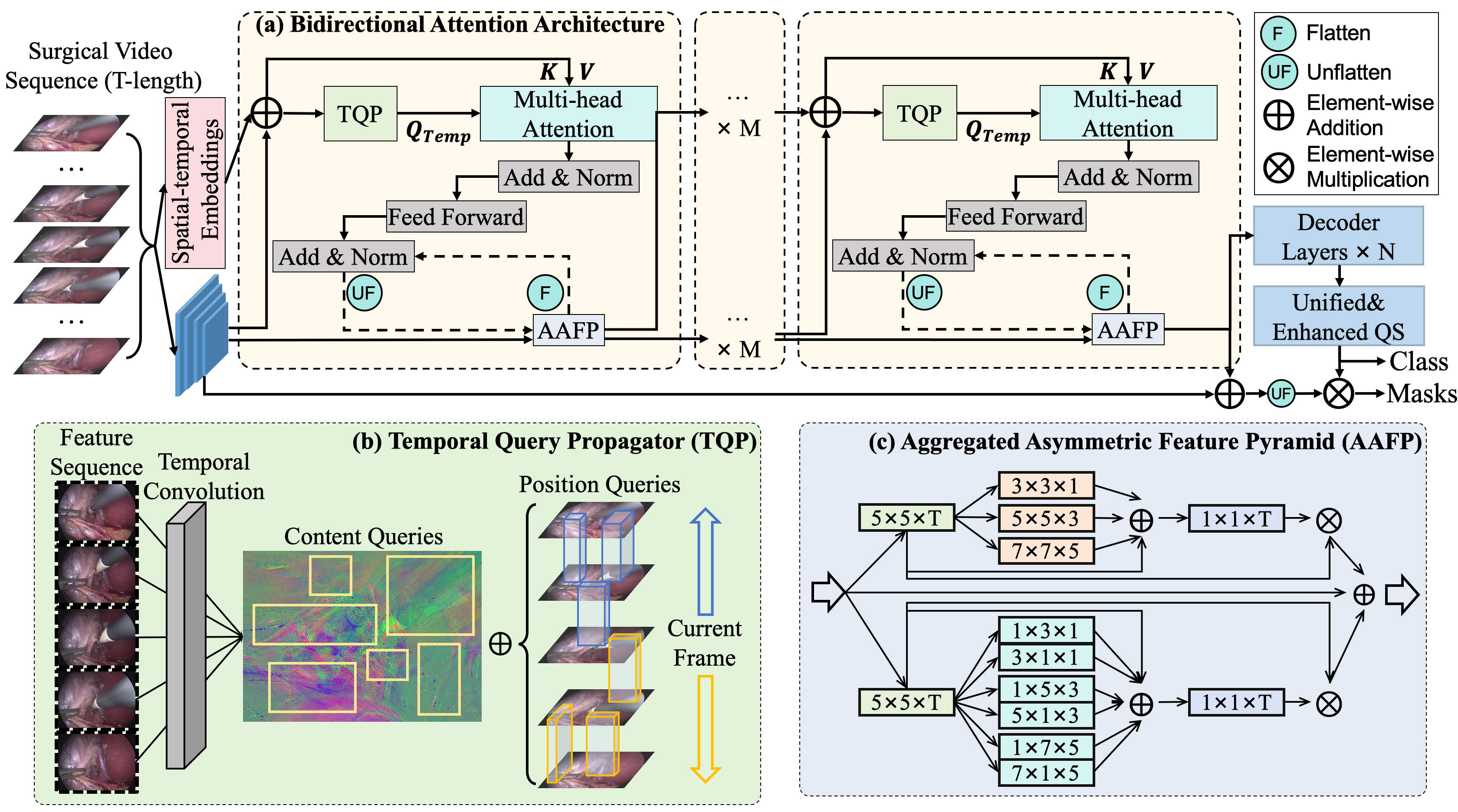}
\caption{The overall framework of TAFPNet. It contains a (a) bidirectional attention architecture injected with the (b) Temporal Query Propagator (TQP) and the (c) Aggregated Asymmetric Feature Pyramid (AAFP) module.} \label{fig2}
\end{figure}

In this section, we present the proposed TAFPNet, which incorporates the Temporal Query Propagator (TQP) and the Aggregated Asymmetric Feature Pyramid (AAFP) to a bidirectional attention network for accurate surgical scene segmentation from laparoscopic surgery videos. The detailed architecture of TAFPNet is illustrated in Fig.~\ref{fig2}.

\subsection{Bidirectional Attention Architecture}
We construct a bidirectional attention architecture to take advantage of progressive feature interaction between a dual-feature extraction branch, a transformer encoding, and a convolution pyramid, to a large extent. Given an image sequence, the multiscale feature pyramid is generated by a stack of 3D convolutions at resolutions of \{1/4, 1/8, 1/16, 1/32\} relative to the input. Subsequently, a multistage feature interaction process is constructed to fuse multi-perceptive information, where each stage comprises two pathways. In the transformer encoding, the input feature embedding passes through the TQP module, multihead attention, and dimensional expansion, then added to the output of the convolution branch. Meanwhile, in the convolution encoding, the input feature pyramid is enhanced by the AAFP module and then compressed into the same dimension embedding to add to the output of transformer branch. After M-stage bidirectional attention, fused features are fed into the transformer decoder for mask prediction. Therefore, it enables the progressive refinement of spatial precision and temporal coherence in surgical video, by implementing parameter-shared residual connection and element-wise summation.

\subsection{Temporal Query Propagator}
The way of incorporating temporal coherence in surgical video analysis is inherently crucial to leverage its benefits. For such purposes, we design a temporal query propagator, namely TQP, which introduces temporal information into the query generation and propagation. This strategy ensures consistent segmentation results across observed frames, so it can generalize well to some cases of rapid instrument manipulation and anatomy-instrument occlusion.

Given a sequence of $T$ length consecutive frames, we first extract frame-wise features ${F_1,F_2,…,F_T}$ by a ResNet-50 backbone. The obtained features are then projected with learnable weights $W^K$ and $W^V$. TQP is an attention module which is further applied to propagate the projected feature along the temporal axis. The key $K$ and the value $V$ of the TQP are calculated based on feature pyramid $\mathrm{concat}\{F_t\}_{t=1}^T$. The temporal query $Q$ is a learned embedding, which is designed to efficiently aggregate the information in regions of interest along the temporal axis. It contains two parts, the content query $Q^{cont}$ and the position query $Q^{pos}$. The content query is calculated by:

\begin{equation}
Q^{content}=\text{top-}K(\mathrm{Conv}\left(\mathrm{concat}\{F_t\}_{t=1}^T\right)).
\end{equation}
where $\text{top-}K(\cdot)$ represents top-k query selection based on the high activation score, and $\mathrm{concat}(\cdot)$ represents concatenation along with time dimension. The position query of the $t$-th frame represents the position embedding of a spatial-temporal tube, which is obtained by:
\begin{equation}
    Q^{pos}_t=
    \begin{cases}
    Linear\left(Q^{pos}_{t+1}\right) &\text{if } 1\leq t\leq(T-1)/2\\
    W^Q\mathrm{Conv}\left(\mathrm{concat}\{F_t\}_{t=1}^T\right)  &\text{if } t=(T+1)/2\\
    Linear\left(Q^{pos}_{t-1}\right) &\text{if } (T+1)/2< t\leq T
    \end{cases}
\end{equation}
where $W^Q$ is a learnable weight. Subsequently, we feed the content query, the position query, the key, and the value into multihead attention layers for feature refinement:
\begin{equation}    
\mathrm{Attention}\left(Q,K,V\right)=\mathrm{Softmax}\left(\frac{(\sum_{t=1}^T{Q_t^{pos})} \bigodot Q^{content}K^T}{\sqrt{d_k}}\right)V 
\end{equation}
where $\bigodot$ represents the dot product operator. After training, each of the selected temporal queries corresponds to a spatial-temporal feature tube, see Fig.~\ref{fig2} (b).  

\subsection{Aggregated Asymmetric Feature Pyramid}
Characteristics of anatomy and instrument generally have obvious differences in visual perspective, presenting irregular-polygon and regular-bar shape, separately. In addition, the same anatomy or instrument maintains inherent topological structure invariance in a video sequence. Thus, we design an aggregated asymmetric Feature Pyramid, namely AAFP, to comprehensively aggregate frame features, from global to local. It contains two perception enhancement operations for anatomy and instruments, separately.

The input feature pyramid $\mathrm{concat}\{F_t\}_{t=1}^T$ first passes through a convolution with the kernel size of $\left(5\times5\times{T}\right)$ to obtain a spatial-temporal aggregating feature map $F_{temp}$. For anatomy perception, we adopt symmetric convolutions to enhance irregular-polygon features:
\begin{equation}
F^{AP}_m=\mathrm{Conv}_{k_m\times{k_m}\times{t_m}}\left(F_{temp}\right)
\end{equation}
where multiple $\left(k_m,k_m,t_m\right)$ pairs are set as $\left(3,3,1\right)$, $\left(5,5,3\right)$, and $\left(7,7,5\right)$. For instrument perception, we adopt asymmetric calculations by parallel strip convolution pairs to enhance the regular-bar features:
\begin{equation}
F^{IP}_m=\mathrm{Conv}_{k_m\times{1}\times{t_m}}\left(F_{temp}\right)+\mathrm{Conv}_{1\times{k_m}\times{t_m}}\left(F_{temp}\right)
\end{equation}
where $k_m$ and $t_m$ are same with them in symmetric convolutions. Subsequently, the final aggregated attention map of registered as $E_{temp}$ is calculated by:
\begin{equation}
E_{temp}=\mathrm{Conv}_{1\times{1}\times{T}}\left(\sum_{m=0}^M{\left(F^P_m\right)+F_{temp}}\right)\bigotimes{F_{temp}}
\end{equation}
where $M$, $\bigotimes$, and $F^{P}_m$ represents the number of selected kernels, matrix multiplication, and the enhanced feature map of anatomy or instruments. Finally, aggregated attention maps from both anatomy and instrument perception blocks are added as the final output.

\section{Experiments and Results}

\subsubsection{Datasets and Evaluation Metrics.} The experiments were performed on two laparoscopic surgery datasets, and our proposed method was conducted a comparative analysis with existing state-of-the-art (SOTA) methods.

The first dataset named EndoVis2018~\cite{ref16} consists of 19 sequences, officially divided into 15 training sets and 4 test sets. Each sequence contains about 250 stereo-pair frames with 1280×1024 resolution. The segmentation ground truth is released by the challenge organizer, and here we only utilize the annotated left image sequences. This segmentation task is defined to divide the entire surgical scene into 12 categories, including different anatomies and instruments.

The second dataset named Endoscapes2023~\cite{ref17} was recently released for three sub-tasks, that are surgical scene segmentation, object detection, and critical view of safety assessment. Here, we use its subset consisting of 493 annotated frames from 50 laparoscopic cholecystectomy videos. It is officially divided into 3:1:1 ratio as the training set, validation set, and test set. Compared to EndoVis2018, this segmentation task is more challenging as it requires the identification of six distinct types of anatomical structures and instruments from the entire image.

To capture temporal clues, we utilize the temporal window to cover 5\% of the mean sequence duration as offline datasets. In the test, we follow the same evaluation manner as the challenge required for a direct and fair comparison. We also use the official evaluation protocol, i.e., (1) mean intersection-over-union (mIoU) and mean Dice coefficient (mDice) for EndoVis2018, and (2) both detection and segmentation mean average precision (mAP@[0.5:0.95]) for Endoscapes2023.

\subsubsection{Implementation Details.}  The resolution of EndoVis2018 and Endoscapes2023 image is reduced to 512×1024 resolution for memory saving. Our proposed method is implemented in PyTorch (2.2.1 version) with 2 NVIDIA A40 GPUs for calculation. Double GPUs only enable the network to be trained in the batch size of 4. Model training iteration and base learning rate are set to 20k and 1e-4, separately. Deeper networks did not produce obvious improvements in early trials, so the backbone of our network is stuck to ResNet-50~\cite{ref18}.

\subsubsection{Comparison with State-of-the-Art Methods.} We compare our TAFPNet with SOTA methods. On EndoVis2018, Table~\ref{table1} lists the compared methods in the first line, involving in three categories: (1) Three reported methods in the 2018 Robot Scene Segmentation Challenge like NCT, UNC, and OTH~\cite{ref16}; (2) Multiscale feature fusion networks such as U-Net~\cite{ref19}, DeepLabv3+~\cite{ref4}, UPerNet~\cite{ref20}, and HRNet~\cite{ref21}; (3) Specific-designed segmentation methods, such as SegFormer~\cite{ref22}, SegNeXt~\cite{ref23}, STswinCL~\cite{ref8}, and LSKANet~\cite{ref6}. Our TAFPNet consistently outperforms in all four test sequences and exceeds the previous SOTA results by 16.4\% at mIoU and 14.6\% at mDice. In particular, the mIoU of sequence 2 and 4 outperforms the previous best performance with a huge improvement of 21.6\% and 31.7\%. Although compared methods constructed advanced spatial feature enhancement strategies, our TAFPNet combines temporal dynamics with structural asymmetry to explicitly generate reliable guidance for segmentation.

\begin{table}
\caption{Quantitative results on EndoVis2018. The best results are bold, and the second best results are underlined.}\label{table1}
\resizebox{\textwidth}{!}{
\begin{tabular}{|c|c|ccccccccc|ccc|}
\hline
\multicolumn{2}{|c|}{Method}&\makecell{OTH~\cite{ref16}}&\makecell{U-Net\\~\cite{ref19}}&\makecell{DeepLab\\v3+~\cite{ref4}}&\makecell{UPer\\Net~\cite{ref20}}&\makecell{HR\\Net~\cite{ref21}}&\makecell{SegFor\\mer\cite{ref22}}&\makecell{SegNe\\Xt~\cite{ref23}}&\makecell{STswin\\CL~\cite{ref8}}&\makecell{LSKA\\Net~\cite{ref6}}&\makecell{Base\\Net}&\makecell{AFP\\Net}&\textbf{\makecell{TAFPNet\\ (Ours)}}\\
\hline
\multirow{5}{*}{mIoU (\%)}  &Seq1&69.1&55.6&64.1&67.4&68.9&69.1&\underline{70.6}&67.0&67.5&54.1&70.5&\textbf{76.2}\\
{}                          &Seq2&57.5&50.5&57.0&54.0&57.3&55.8&57.1&63.4&63.2&63.8&\underline{83.1}&\textbf{85.4}\\
{}                          &Seq3&82.9&69.7&82.3&82.0&85.0&81.6&84.3&83.7&\underline{85.2}&70.8&83.8&\textbf{88.1}\\
{}                          &Seq4&39.0&26.8&31.6&30.2&42.1&45.5&45.1&40.3&48.9&47.0&\underline{72.4}&\textbf{80.6}\\
{}                          &Overall&62.1&50.7&58.8&58.4&63.3&63.0&64.3&63.6&66.2&58.9&\underline{77.5}&\textbf{82.6}\\
\hline
\multicolumn{2}{|c|}{mDice (\%)}& --& 61.5&67.3&66.8&71.8&71.9&72.5&72.0&75.3&69.3&\underline{86.6}&\textbf{89.9}\\
\hline
\end{tabular}
}
\end{table}

Table~\ref{table2} shows the performance of compared methods on Endoscapes2023. The compared methods consist of three fine-tuned segmentation models, namely Mask-RCNN~\cite{ref24}, Cascade Mask-RCNN~\cite{ref25}, and Mask2Former~\cite{ref9}. Although the detection and segmentation performance of each category differs greatly in the compared methods, our TAFPNet achieves the best overall results with the detection mAP of 30.8\% and the segmentation mAP of 29.9\%. However, a substantial performance gap persists in fine-grained object recognition, revealing the extra difficulties of Endoscapes2023, probably due to worse lighting situation and more motion blurs.

\begin{table}
\caption{Quantitative results on Endoscapes2023. The best results are bold, and the second best results are underlined.}\label{table2}
\resizebox{\textwidth}{!}{
\begin{tabular}{|c|ccccccc|}
\hline
\multirow{2}{*}{Method}& \multicolumn{7}{|c|}{Detection mAP@[0.5:0.95] $\mid$ Segmentation mAP@[0.5:0.95] (\%)}\\
    \cline{2-8}
    {}&  Cystic Plate&  HC Triangle&  Cystic Artery&  Cystic Duct&  Gallbladder&  Tool&  Overall\\
    \hline
    Mask-RCNN~\cite{ref24}& 2.8 $\mid$ 3.3&  2.9 $\mid$ 3.8&  12.7 $\mid$ 11.9&  7.4 $\mid$ 7.9&  45.8 $\mid$ 59.1& 49.7 $\mid$ 51.2& 20.2 $\mid$ 22.9\\ 
    Cascaded Mask-RCNN~\cite{ref25}& \underline{3.1} $\mid$ \textbf{6.5}&  11.1 $\mid$ 6.7&  \underline{15.7} $\mid$ 10.8& 11.7 $\mid$ 10.0&  62.0 $\mid$ 62.7&  63.2 $\mid$ 56.8& 27.8 $\mid$ 25.6\\
    Mask2Former~\cite{ref9}&  1.4 $\mid$ 1.7&  \underline{14.3} $\mid$ 8.3&  6.4 $\mid$ 7.6&  \underline{14.7} $\mid$ \textbf{15.9}&  \textbf{68.7} $\mid$ 62.6&  67.1 $\mid$ \textbf{63.5}&  28.8 $\mid$ 26.6\\ 
    \hline
    BaseNet&  1.1 $\mid$ 0.5&  12.8 $\mid$ 10.4&  9.8 $\mid$ 12.2&  7.2 $\mid$ 7.9&  64.0 $\mid$ \underline{63.2}&  58.7 $\mid$ 59.5&  25.6 $\mid$ 25.6\\ 
    AFPNet&  2.8 $\mid$ 3.0&  5.6 $\mid$ \underline{10.7}&  \textbf{16.5} $\mid$ \textbf{20.1}&  \textbf{14.9} $\mid$ 10.1& \underline{67.1} $\mid$ \textbf{65.4} &  \textbf{72.2} $\mid$ 62.2&  \underline{29.8} $\mid$ \underline{28.7}\\ 
    \textbf{TAFPNet (Ours)}& \textbf{10.5} $\mid$ \underline{6.3}& \textbf{16.3} $\mid$ \textbf{23.1}& 14.9 $\mid$ \underline{16.4}& 10.6 $\mid$ \underline{10.3}& 63.5 $\mid$ 60.5& \underline{68.9} $\mid$ \underline{62.8}& \textbf{30.8} $\mid$ \textbf{29.9}\\
    \hline
\end{tabular}
}
\end{table}

\begin{figure}
\includegraphics[width=\textwidth]{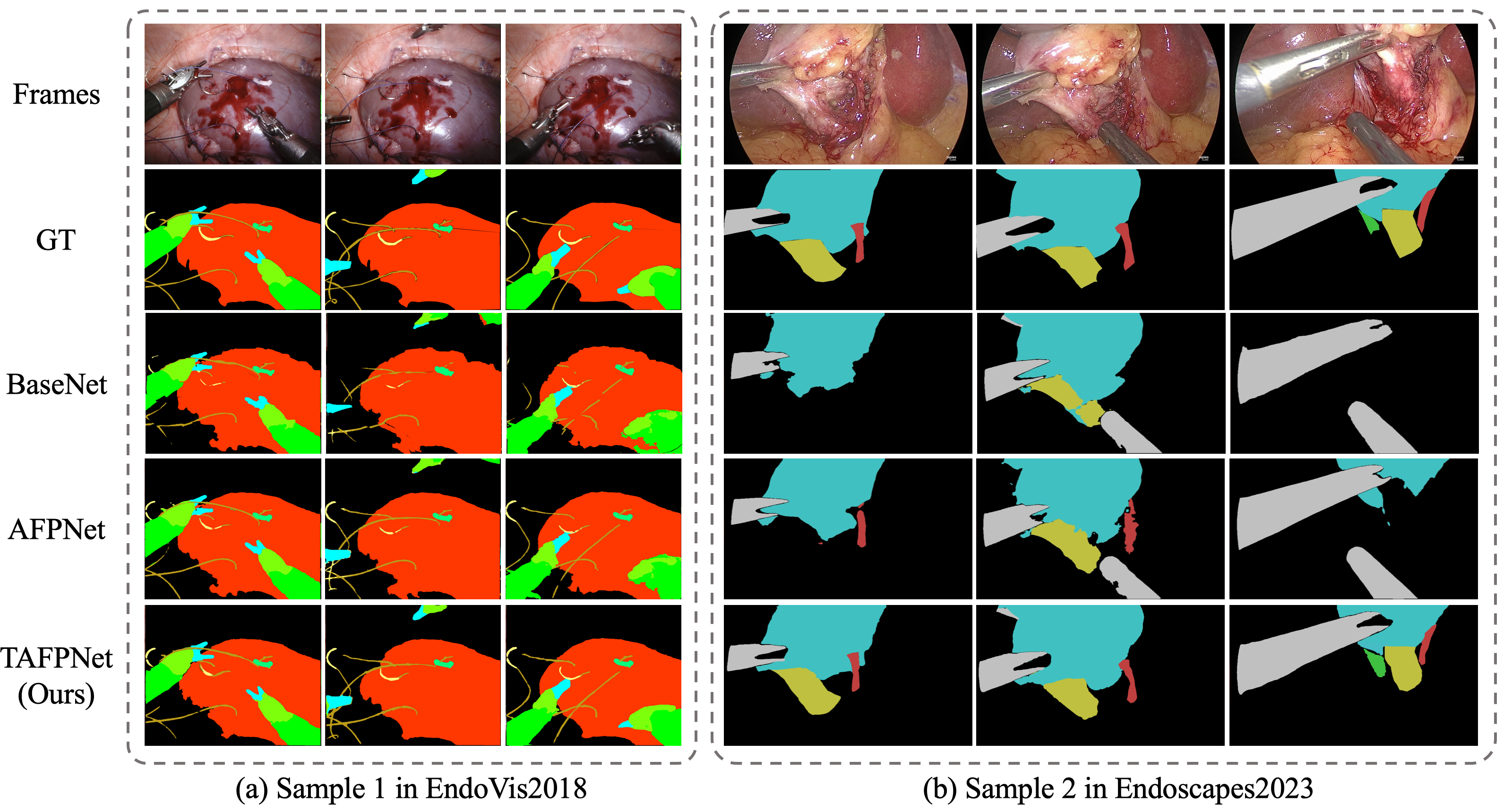}
\caption{Visual comparison of segmentation results on (a) EndoVis2018 and (b) Endoscapes2023. From top to bottom, for each dataset, we present three continuous video frames and their corresponding ground truth, with segmentation results using BaseNet, AFPNet and our proposed TAFPNet.} \label{fig3}
\end{figure}

\subsubsection{Effectiveness of Temporal Propagation and Asymmetric Attention.} We evaluate the effectiveness of two core enhancement modules in our TAFPNet. The results of three ablation settings are also listed in Table~\ref{table1} and Table~\ref{table2}: (1) a plain vision transformer encoder-decoder network as the baseline (BaseNet), (2) an Asymmetric Feature Propagation Network (AFPNet) without temporal coherence guidance, and (3) our proposed framework (TAFPNet). We adopt the same backbone for different settings for fairness. It can be seen that AFPNet performs better than BaseNet, especially on EndoVis2018, demonstrating the effectiveness of the bidirectional attention architecture. Furthermore, our proposed TAFPNet further increases performance in almost all categories, except for the cystic artery and gallbladder in Endoscapes2023. Although dim ambient light causes low recognition precision scores on Endoscapes2023, our proposed TAFPNet still identifies delicate structures such as the cystic artery and the cystic plate, which are terribly missed in compared ablation networks. This demonstrates that temporal propagation significantly enhances feature propagating precision. Fig~\ref{fig3} exhibits some visual results of ablation experiments. TAFPNet can achieve the complete and consecutive segmentation of slender thread in sequence 1, which is misidentified in other ablation networks. Besides, the blurred visual boundary of rapidly moved clasper in sequence 2 is more accurately and clearly recognized by our TAFPNet.In summary, our proposed TAFPNet shows a noticeable advancement in the ability of fine-grained structure recognition and fast motion perception.

\section{Conclusion}
In this work, we propose a novel interaction framework for surgical scene segmentation that incorporates temporal dependency in query propagation and disentangled feature representation. Our method achieves the best overall performance in both scene segmentation on EndoVis2018 and object recognition on Endoscapes2023, especially on fine-grained structures. The superior performance establishes a promising value for clinical robot-assisted intervention.

\end{document}